\documentclass[conference,fleqn]{IEEEtran}
\usepackage{bbm, balance}
\usepackage[noend]{algpseudocode}
\usepackage{blindtext, graphicx}
\usepackage {tikz}
\usepackage{caption}
\usepackage{amssymb}
\usepackage{float}
\usepackage{algpseudocode}
\usetikzlibrary {positioning}
\usepackage{tkz-euclide}
\usepackage{algorithm,algcompatible,amsmath}
\usetkzobj{all}
\usepackage[T1]{fontenc}
\usepackage[bookmarks=false]{hyperref}
\usepackage{url}
\usepackage[utf8x]{inputenc}
\usetikzlibrary{calc}
\usepackage{amsthm}
\usepackage{thmtools}
\usepackage[algo2e]{algorithm2e} 
\definecolor {processblue}{cmyk}{0.96,0,0,0}
\ifCLASSINFOpdf
 
\else

\fi

\newcommand{\Lagr}{\mathcal{L}}
\begin{document}

\title{The Effect of Communication on Noncooperative Multiplayer Multi-Armed Bandit Problems}

\author{
        \IEEEauthorblockN{Noyan Evirgen\IEEEauthorrefmark{1}, Alper Köse\IEEEauthorrefmark{2}\IEEEauthorrefmark{3}}
        \IEEEauthorblockA{
                \small
                \IEEEauthorrefmark{1}Department of Information Technology and Electrical Engineering, ETH Zurich \\
                \IEEEauthorrefmark{2}Research Laboratory of Electronics, Massachusetts Institute of Technology \\       
                \IEEEauthorrefmark{3}Department of Electrical Engineering, École Polytechnique Fédérale de Lausanne \\
                \vspace{-0.5cm}
        }
    
}

\maketitle
\footnotetext[1]{This work has been accepted to the 2017 IEEE ICMLA.}

\begin{abstract}

We consider decentralized stochastic multi-armed bandit problem with multiple players in the case of different communication probabilities between players. Each player makes a decision of pulling an arm without cooperation while aiming to maximize his or her reward but informs his or her neighbors in the end of every turn about the arm he or she pulled and the reward he or she got. Neighbors of players are determined according to an Erd{\H{o}}s-R{\'e}nyi graph with  connectivity $\alpha$ which is reproduced in the beginning of every turn. We consider i.i.d. rewards generated by a Bernoulli distribution and assume that players are unaware about the arms' probability distributions and their mean values. In case of a collision, we assume that only one of the players who is randomly chosen gets the reward where the others get zero reward. We study the effects of $\alpha$, the degree of communication between players, on the cumulative regret using well-known algorithms UCB1, $\epsilon$-Greedy and Thompson Sampling.

\end{abstract}


\IEEEpeerreviewmaketitle

\section{Introduction}

In Multi-armed Bandit (MAB) problem, players are asked to choose an arm which returns a reward according to a probability distribution. In MAB, we face an exploration-exploitation trade-off. Exploration can be interpreted as a search for the best arm while exploitation can be thought as maximizing reward or minimizing regret by pulling the best arm. Therefore, we must search enough to be nearly sure that we find the best arm without sacrificing much from the reward. There are different kinds of MAB problems that can be studied:

\begin{itemize}

\item \textbf{Stochastic MAB}: Each arm $i$ has a probability distribution $p_i$ on [0,1], and rewards of arm $i$ are drawn i.i.d. from $p_i$ where distribution $p_i$ does not change according to the decisions of a player. In \cite{auer2002finite}, stochastic MAB setting can be seen.

\item \textbf{Adversarial MAB}: No statistical assumptions are made on the rewards. In \cite{auer1995gambling}, authors give a solution to the adversarial MAB.

\item \textbf{Markovian MAB}: Each arm $i$ changes its state as in a markov chain when it is pulled and rewards are given depending on the state. In \cite{tekin2010online}, the classical MAB problem with Markovian rewards is evaluated.

\end{itemize}

MAB problem is introduced by Robbins \cite{robbins1952some} and investigated under many different conditions. Auer et al. \cite{auer2002finite} show some of the basic algorithms in a single player model where the considered performance metric is the regret of the decisions. Koc{\'a}k et al. \cite{kocak2016online} consider adversarial MAB problems where player is allowed to observe losses of a number of arms beside the arm that he or she actually chose and each non-chosen arm reveals its loss with an unknown probability. Kalathil et al. \cite{kalathil2014decentralized} consider decentralized MAB problem with multiple players where no communication is assumed between players. Also, arms give different rewards to different players and in case of a collision, no one gets the reward. Liu and Zhao \cite{liu2010distributed} compare multiple players without communication and multiple players acting as a single entity scenarios where reward is assumed to be shared in an arbitrary way in case of a collision. MAB can be used in different type of applications including cognitive radio networks and radio spectrum management as seen in \cite{lai2008medium}, \cite{gai2011combinatorial} and \cite{anandkumar2010opportunistic}.

In this paper, we study a decentralized MAB, and consider the scenario as $N<S$ where $N$ denotes the number of players and $S$ denotes the number of arms. Players exchange information in the end of every turn according to Erd{\H{o}}s-R{\'e}nyi communication graph which is randomly reproduced every turn with $\alpha$ connectivity, $ 0\leq \alpha \leq 1$. Also, we consider collisions in our scenario, where only one randomly chosen player gets the reward where other players get zero reward. Our goal is to minimize the cumulative regret in the model where all players use the same algorithm while making their decisions. To this end, we use three different well-known MAB algorithms, Thompson Sampling \cite{agrawal2012analysis}, $\epsilon$-Greedy \cite{auer2002finite} and UCB1 \cite{auer2002finite}. In the considered scenario, everybody is alone in the sense that all players make decisions themselves, and everybody works together in the sense that there can be a communication between players in the end of every turn. 

The paper is organized as follows. We formulate the problem in Section II. We explain our reasoning and propose optimal policies in case of $\alpha=0$ and $\alpha=1$ in Section III. Then, we discuss the simulation results where we have $Cumulative$ $Regret$ $vs$ $\alpha$ graph and $Cumulative$ $Regret$ $vs$ $Number$ $of$ $Turns$ graphs and we have these results for two different mean distributions of arms in Section IV. Finally, we conclude our findings in Section V.

\section{Problem Formulation}

We consider a decentralized MAB problem with $N$ players and $S$ arms. In our model, players are allowed to communicate according to an Erd\H{o}s-Renyi random graph with connectivity $\alpha$, so each player $p$ informs its neighbours $N(p)$ about the arm it pulled and the reward it earned in the end of each turn. In other words, let us think a system graph $\mathcal{G}=(\mathcal{P}, \mathcal{E})$. Players are shown as vertices, $p_{k}\in \mathcal{P}$ where $k=1...N$ and $\{p_{a},p_{b}\}\in \mathcal{E}$ if there is a connection between players $p_{a}$ and $p_{b}$, which is true with probability $\alpha$.

One turn is defined as a time interval in which every player pulls an arm according to their game strategy. Note that the random communication graph changes every turn but $\alpha$ is constant. 

In addition to the aforementioned setup, each arm yields a reward with a random variable $X_{i,t}$ associated to it, where $i$ is the index of an arm and $t$ is the turn number. Successive pulls of an arm are independent and identically distributed according to a Bernoulli distribution with expected value of $\mu_i$, which are unknown to the players.

Because of the nature of the problem, "collision" should also be considered. When an arm with index $i$ is chosen by multiple players, only one of the players, chosen randomly, receive the reward $X_{i,t}$ whereas the rest of the players receives zero reward. Players are not aware of the collision model.

We can define the expected cumulative regret in a single player model as:

\begin{equation} 
R_{p,T}=T\max_{i \in 1...S}\mu_i - \sum_{k=1}^{T} \mu_{Y_{p,k}}
\end{equation}

where $Y_{p,k}$ is the chosen arm index in the $k$th turn of pulls by the player $p$. However, for our model having multiple players, we do not want all players to go for the best arm due to collision model. That is to say, in our setting players affect each other's reward. Therefore, we cannot define the regret per player and independently sum them, instead we directly define the cumulative regret in the game based on the total expected reward. The cumulative regret in the game can be defined as: 

\begin{equation} 
R_{T} = T\max_{\forall a_p \in \{1...S\}, i\neq j \Rightarrow a_i\neq a_j} (\sum_{p=1}^{N} \mu_{a_p}) - \sum_{p=1}^{N}\sum_{k=1}^{T} \mu_{Y_{p,k}}
\end{equation}

where $a_i$ is the index of hypothetically chosen slot by the $i$th player. Since the first term of the right hand side is a constant, it can be seen that the strategy which minimizes the cumulative regret is the one which maximizes $\sum_{p=1}^{N}\sum_{k=1}^{T} \mu_{Y_{p,k}}$. Minimizing cumulative regret adds up to same thing with maximizing total cumulative reward in the game. Because of the collision model, total cumulative reward does not depend on the individual pulls. Instead, it can be calculated based on whether an arm is chosen at a certain turn. Therefore, total cumulative reward can be defined as:
\begin{equation}
\label{eq:gain}
G = \sum_{k=1}^{T}\sum_{i=1}^{S}I_{i,k} X_{i,k}
\end{equation}
where $I_{i,k}$ is indicator of whether the arm with index i is chosen at the $k$th turn of pulls. Let us define $\mathbbm{1}\{\cdot\}$ to be the indicator function. Then $I_{i,k}$ can be calculated as:
\begin{equation}
\label{eq:indic}
I_{i,k} = \mathbbm{1}\Bigl\{\Bigl[\sum_{p=1}^{N}\mathbbm{1}\{Y_{p,k} = i\}\Bigr] \neq 0\Bigr\}
\end{equation}
where again $Y_{p,k}$ is the chosen arm index by player p in the $k$th turn of pulls.

\section{System Model}
An important evaluation of strategies is the expected total cumulative reward under the constraints of the problem. Considering players cannot collaboratively plan for their next strategy, it has to be assumed that each player tries to maximize its own reward. The strategy which maximizes the total cumulative reward is the one which assigns N players to different arms which have the highest N expected rewards. Let us define $q_k$ as the $k$th best arm. Then, the expected maximum total cumulative reward after T turns for $\alpha = 0$ can be defined as:

\begin{equation}
R_{max_{T,\alpha = 0}} = T\sum_{k=1}^N\mu_{q_{k}}
\end{equation}

This, combined with the connectivity parameter $\alpha$ introduces an interesting trade-off phenomenon. In order to elaborate this, consider the case where $\alpha = 0$. When there is no communication between the players, each player can converge to a different arm believing their choice is the best one, which is mainly caused by the collision model. Converging here means choosing the same arm after a limited turn of pulls. 

Now consider when $\alpha = 1$ where every player knows everything about other pulls. Inevitably, this results in same probabilistic distributions for every arm for every player. In other words, players cannot converge to different arms. They can either converge to the same arm or not converge at all. Since our reward depends on $I_{i,k}$ from Equation (\ref{eq:indic}), not converging has a higher total cumulative reward than every player converging to the best arm which would only have the reward of that arm. Therefore, the expected maximum total reward when $\alpha = 1$ is when every player randomly chooses an arm with a probability which depends on expected means of the arms, assuming $S > N$. 

In Equation (\ref{eq:gain}), we introduce total cumulative reward which we try to maximize. Let us define a different metric called $L$ which stands for total cumulative loss:

\begin{equation}
\begin{split}
 L & = \sum_{k=1}^{T}\sum_{i=1}^{S}\mathbbm{1}\Bigl\{\Bigl[\sum_{p=1}^{N}\mathbbm{1}\{Y_{p,k} = i\}\Bigr] = 0\Bigr\} X_{i,k}
\\
& = \sum_{k=1}^{T}\sum_{i=1}^{S}\Bigl[1-\mathbbm{1}\Bigl\{\Bigl[\sum_{p=1}^{N}\mathbbm{1}\{Y_{p,k} = i\}\Bigr] \neq 0\Bigr\}\Bigr] X_{i,k}
\\
& = \sum_{k=1}^{T}\sum_{i=1}^{S}X_{i,k} - G
\end{split}
\end{equation}

First term of the right hand side is a constant. Therefore, maximizing $G$ will minimize $L$. Therefore, $\mathbf{E}[L]$ can be minimized if the expected loss of a turn is minimized:
\begin{equation}
\mathbf{E}[L_T] = \sum_{i=1}^{S}\mathbbm{1}\Bigl\{\Bigl[\sum_{p=1}^{N}\mathbbm{1}\{Y_{p} = i\}\Bigr] = 0\Bigr\}\mu_i
\end{equation}

where $Y_{p}$ is the chosen arm index by the player p. Assuming $S > N$ with $S$ number of arms and $N$ players, let us define $c_i$ as the probability of a player choosing arm with index i. Note that, $c_i$ is the same for every player since $\alpha$ equals to 1. Then it can be seen that, $\sum_{i=1}^{S}c_i = 1$. Therefore the expected loss of a turn can be defined as:
\begin{equation}
\begin{split}
\mathbf{E}[L_T] & = \sum_{i=1}^{S}\mathbbm{1}\Bigl\{\Bigl[\sum_{p=1}^{N}\mathbbm{1}\{Y_{p} = i\}\Bigr] = 0\Bigr\}\mu_i \\
& = \sum_{i=1}^{S}(1-c_i)^N\mu_i = \sum_{i=1}^{S}m_i^N\mu_i
\end{split}
\end{equation}

where $m_i$ is $1-c_i$. Note that $0 \leq m_i \leq 1$. In order to minimize expected loss of a turn, the Lagrangian which we try to maximize can be defined as:
\begin{equation}
\Lagr(m_i,\lambda_i) = -\sum_{i=1}^{S}\Bigl[m_i^N\mu_i  + \lambda_{2i-1}m_i + \lambda_{2i}(1-m_i)\Bigr] 
\end{equation}
Since,
\begin{equation}
\label{eq:eseksi}
\begin{split}
&\sum_{i=1}^{S}c_i = 1 \\
&\sum_{i=1}^{S}m_i = \sum_{i=1}^{S}(1-c_i) = S - 1\\
\Rightarrow&\frac{\partial m_x}{\partial m_{i \neq x}} = -1 
\end{split}
\end{equation}
where $1 \leq x \leq S$. Then in order to maximize the Lagrangian,
\begin{equation}
\begin{split}
\frac{\partial \Lagr}{\partial m_x} &=  -N(m_x^{N-1}\mu_x-\sum_{i = 1, i \neq x}^S m_i^{N-1}\mu_i) \\
& + \lambda_{2x-1}-\lambda_{2x} + \sum_{i=1,i \neq x}^S\lambda_{2i} - \sum_{i=1,i \neq x}^S\lambda_{2i-1} = 0  
\end{split}
\end{equation}

From Karush-Kuhn-Tucker conditions (KKT), $\lambda_{2i-1}m_i = 0$, $\lambda_{2i}(1-m_i) = 0$. $m_i = 1$ is a case where the players do not pull the arm with index i. Similar with $m_i = 0$, where players only pull the arm with index i. Both of these cases can be ignored if there is a valid solution without them. Otherwise, $m_i = 1$ case will be revisited starting from the machine with the lowest expected mean $\mu_i$. For the derivation of the solution let us assume $\lambda_{2i-1} = \lambda_{2i} = 0$. Then,

\begin{equation}
\begin{split}
\frac{\partial \Lagr}{\partial m_{x_1}} &= -N(m_{x_1}^{N-1}\mu_{x_1}-\sum_{i = 1, i \neq x_1}^S m_i^{N-1}\mu_i) = 0\\
\frac{\partial \Lagr}{\partial m_{x_2}} &= -N(m_{x_2}^{N-1}\mu_{x_2}-\sum_{i = 1, i \neq x_2}^S m_i^{N-1}\mu_i) = 0
\end{split}
\end{equation}

where $x_1 \neq x_2$, $1 \leq x_1 \leq S$ and $1 \leq x_2 \leq S$. Therefore,
\begin{equation}
\begin{split}
m_{x_1}^{N-1}\mu_{x_1}-\sum_{i = 1, i \neq x_1}^S &m_i^{N-1}\mu_i = \\
&m_{x_2}^{N-1}\mu_{x_2}-\sum_{i = 1, i \neq x_2}^S m_i^{N-1}\mu_i
\end{split}
\end{equation}
\begin{equation}
\begin{split}
m_{x_1}^{N-1}\mu_{x_1}-&m_{x_2}^{N-1}\mu_{x_2} -\sum_{i = 1, i \neq \{x_1,x_2\}}^S m_i^{N-1}\mu_i =\\ &m_{x_2}^{N-1}\mu_{x_2}-m_{x_1}^{N-1}\mu_{x_1} -\sum_{i = 1, i \neq \{x_1,x_2\}}^S m_i^{N-1}\mu_i \\
m_{x_1}^{N-1}\mu_{x_1} &= m_{x_2}^{N-1}\mu_{x_2}
\end{split}
\end{equation}

Let us assume,

\begin{equation}
\begin{split}
&A  = m_i^{N-1}\mu_i = (1-c_i)^{N-1}\mu_i, \forall i \in \{1...S\} \\
&c_i  = 1 - \sqrt[N-1]{\frac{A}{\mu_i}}\qquad \\
&\sum_{i=1}^{S}c_i = S - \sum_{i=1}^{S}\sqrt[N-1]{\frac{A}{\mu_i}}\qquad = 1 \\
&A  =\Bigg[\dfrac{S-1}{\sum_{i=1}^{S}\sqrt[N-1]{\frac{1}{\mu_i}}\qquad}\Bigg]^{N-1} \\
&c_i  =1 - \dfrac{\Bigg[\dfrac{S-1}{\sum_{k=1}^{S}\sqrt[N-1]{\frac{1}{\mu_k}}\qquad}\Bigg]}{\sqrt[N-1]{\mu_i}\qquad}
\end{split}
\end{equation}

This results in the optimal $c_i$ for the case of $\alpha = 1$ assuming $0 \leq c_i \leq 1$, $\forall i \in {1,2,...,S}$. If the assumed constraint is not satisfied, it means that $\lambda_{2i} \neq 0$ or $\lambda_{2i-1} \neq 0$. For $\lambda_{2i} \neq 0$, since $\lambda_{2i}(1-m_i) = 0$, it means $m_i = 1$ and $c_i = 0$. This conclusion intuitively makes sense; if expected mean of an arm is small enough to force the $c_i$ to become negative, the optimal strategy would be to not pull the arm at all. For $\lambda_{2i-1} \neq 0$, since $\lambda_{2i-1}(m_i) = 0$, therefore $m_i = 0$ and $c_i = 1$. This means that, every player chooses the $i$th arm which is never the optimal play unless the rest of the arms have zero reward. Using these derivations, we introduce an algorithm called asymptotically optimal algorithm which gives an asymptotically optimal strategy for $\alpha = 0$. The algorithm leverages a simulated annealing approach where it either randomly pulls an arm to explore or calculate the optimal $c_i$s to exploit. $c_i$s are then used to sample the arm pull.

Since players are not aware of the collision model, their observed mean estimation for the arms are calculated with the rewards from their neighbors combined with their reward.

\begin{table}
{\LinesNumberedHidden
    \begin{algorithm}[H]
        \SetKwInOut{Input}{Input}
        \SetKwInOut{Output}{Output}
        \SetAlgorithmName{Algorithm}{}
        
        $S$ is the arm count.\\
        $N$ is the player count.\\
        $\mu'_i$ is the observed mean reward of arm with index $i$ for the current player.\\
        For $0 < k < 1$.\\
        \For {t = 1,2,...}{
        $random$ = Random a value between 0 and 1.\\ 
            \If{$1-\epsilon > random$}{
                $\mathcal{H} = \{1,2,...S\}$.\\
                $c_i = 0, \quad \forall i \in \mathcal{H}$.\\
                \While{$\exists i\in\mathcal{H}$ with $(c_i \leq 0) $} {
                    \For {i = 1,2,...,S}{
                        \If{$i \in \mathcal{H}$} {
                            $c_i =1 - \dfrac{\Bigg[\dfrac{S-1}{\sum_{k \in \mathcal{H}}\sqrt[N-1]{\frac{1}{\mu_k'}}\qquad}\Bigg]}{\sqrt[N-1]{\mu_i'}\qquad}$.\\
                            \If {$c_i \leq 0 $} {
                                Discard $i$ from $\mathcal{H}$.
                            }
                        }
                    }
                    
                    \For {i = 1,2,...,S}{
                        \If{$i \in \mathcal{H}$} {
                            $c_i =1 - \dfrac{\Bigg[\dfrac{S-1}{\sum_{k \in \mathcal{H}}\sqrt[N-1]{\frac{1}{\mu_k'}}\qquad}\Bigg]}{\sqrt[N-1]{\mu_i'}\qquad}$.\\
                        }
                    }
                }
                $random$ = Random a value between 0 and 1.\\
                $sum\_of\_chances = 0$.\\
                \For{$i \in \mathcal{H}$}{
                    $sum\_of\_chances += c_i$.\\
                    \If{$sum\_of\_chances \geq random$}{
                        Pull $i$th arm.
                    }
                }
            }
            \Else{
                Randomly pull an arm.\\
            }   
            $\epsilon = \epsilon * k.$
        }  
\caption{Asymptotically Optimal Algorithm for $\alpha=1$ case}
\end{algorithm}}
\caption{Asymptotically Optimal Algorithm for $\alpha=1$ case}
\label{algo1}
\end{table}

\section{Simulation Results}

We do six different simulations to see the effect of communication in MAB problem. In the setup of all simulations, we set $S=10$ and $N=5$. On the other hand, $\mu$ vector has two different value sets, where $\mu_1=[0.9, 0.8, 0.7, 0.6, 0.5, 0.4, 0.3, 0.2, 0.1, 0.01]$ and $\mu_2=[0.7, 0.68, 0.66, 0.64, 0.62, 0.4, 0.38, 0.36, 0.34, 0.32]$. We evaluate the effect of connectivity $\alpha$ for three different algorithms and also propose asymptotic limits for total cumulative reward for $\alpha=0$ and $\alpha=1$ cases, which mean no communication and full communication, respectively. In general, cumulative regret increases with increasing $\alpha$. We get the best results for $\alpha=0$, which means there is no communication between players. This is exactly as we expected due to the collision model we use and can be explained by players' disinclination to pull the same arm due to their different estimations on the means of the arms. Therefore, each player tends to pull a different arm which maximizes the reward. On the other hand, in $\alpha=1$ case, all players have the same mean updates for the arms and they behave similarly. So, when there is an arm with high mean $\mu_i$, all of the players are more inclined to pull this arm, which eventually decreases $\mu_i$ due to collisions. In the end, this forms a balance which makes the probability of pulling each arm similar. This causes a higher probability of collision compared to $\alpha=0$ case and decreases the cumulative reward in the system. We test three well-known algorithms of MAB problem which are modified for communications between players. The aim is to understand how robust are these algorithms against communication between players. $\epsilon$-Greedy and UCB1 can be considered as nearly deterministic algorithms which makes them inevitably fail against communication. Interestingly, they could still provide decent total cumulative rewards until $\alpha = 0.9$. This is mostly caused by their "greedy" nature; even though the observed means for arms are close to each other, players using these algorithm choose the best option. This greediness pays off since players can experience different means even with high amount of connection which results in convergence to different arms. On the other hand, Thompson Sampling is a probabilistic approach. Thus, when players have similar means they choose an arm with similar probabilities which results in lower total cumulative reward for high $\alpha$. However, because of the probabilistic nature of the algorithm, it never catastrophically fails. 

\begin{table}[H]
{\LinesNumberedHidden
    \begin{algorithm}[H]
        \SetKwInOut{Input}{Input}
        \SetKwInOut{Output}{Output}
        \SetAlgorithmName{Algorithm}{}
        
        $\mu'_i$ is the observed mean reward of arm with index $i$ for the current player.\\
        Pull each arm once.\\
        \For {t = 1,2,...}{
            Pull the arm $i(t) = argmax_{i} \mu'_i + \sqrt{\frac{2\ln(n)}{n_{i}}}$ where
            $n_{i}$ is the number of pulls of arm with index $i$ observed by the current player so far and $n$ is the number of arm pulls observed by the current player so far.
        }  
\caption{UCB1 Algorithm \cite{auer2002finite}}
\end{algorithm}}
\caption{UCB1 Algorithm}
\label{algo:UCB1 Algorithm}
\end{table}

\begin{table}[H]
{\LinesNumberedHidden
    \begin{algorithm}[H]
        \SetKwInOut{Input}{Input}
        \SetKwInOut{Output}{Output}
        \SetAlgorithmName{Algorithm}{}

        $\epsilon=1$ and $0<k<1$.\\
        Pull each arm once.\\
        \For {t = 1,2,...}{
            With probability $1-\epsilon$, pull the arm with index $i$ which has the highest mean reward observed by the current player, else pull a random arm.\\
            $\epsilon = \epsilon * k$.            
        }  
\caption{$\epsilon$-Greedy Algorithm \cite{auer2002finite}}
\end{algorithm}}
\caption{$\epsilon$-Greedy Algorithm}
\label{algo:Greedy Algorithm}
\end{table}

\begin{table}[H]
{\LinesNumberedHidden
    \begin{algorithm}[H]
        \SetKwInOut{Input}{Input}
        \SetKwInOut{Output}{Output}
        \SetAlgorithmName{Algorithm}{}
        
        For each arm $i = 1,...,S$ set $S_{i}(1) = 0, F_{i}(1) = 0$.\\
        Pull each arm once.\\
        \For {t = 1,2,...}{
            For each arm $i=1,...,S$, sample $\theta_{i}(t)$ from the $Beta(S_{i} +1,F_{i} +1)$ distribution.\\
            Pull the arm $i(t) = argmax_{i} \theta_{i}(t)$ and observe reward $r$.\\
            If $r= 1$, then $S_{i}(t) = S_{i}(t) + 1$, else $F_{i}(t) = F_{i}(t) + 1$.
        }  
\caption{Thompson Sampling Algorithm \cite{agrawal2012analysis}}
\end{algorithm}}
\caption{Thompson Sampling Algorithm}
\label{algo:Thompson Sampling Algorithm}
\end{table}

As seen in Fig. 3 and Fig. 6, in the full communication scenario, $\epsilon$-Greedy and UCB1 algorithms clearly fail while Thompson Sampling performs nearly as good as the asymptotically optimal method. As seen in Fig. 2 and Fig. 5, in no communication setting, Thompson Sampling and $\epsilon$-Greedy with a good tuned $\epsilon$ perform nearly optimal. On the other hand, Fig. 1 and Fig. 4 show that Thompson Sampling underperforms for other values of $\alpha$. UCB1 and $\epsilon$-Greedy clearly have a lower cumulative regret for $0.5\leq \alpha \leq0.9$.

\begin{figure}[H]
  \centering
    \includegraphics[width=0.45\textwidth]{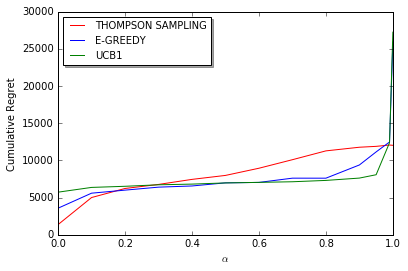}    
    \caption{Change of Cumulative Regret with respect to $\alpha$ where $S=10$, $N=5$ and $\mu=\mu_1$}
\end{figure}

\begin{figure}[H]
  \centering
    \includegraphics[width=0.45\textwidth]{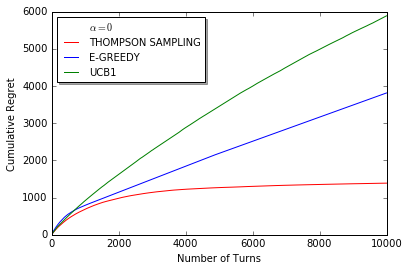}    
    \caption{Change of Cumulative Regret with respect to Number of Turns where $S=10$, $N=5$, $\alpha=0$ and $\mu=\mu_1$}
\end{figure}

\begin{figure}[H]
  \centering
    \includegraphics[width=0.45\textwidth]{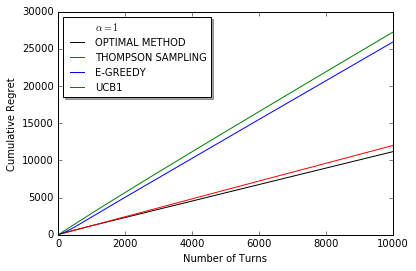}    
    \caption{Change of Cumulative Regret with respect to Number of Turns where $S=10$, $N=5$, $\alpha=1$ and $\mu=\mu_1$}
\end{figure}

\begin{figure}[H]
  \centering
    \includegraphics[width=0.45\textwidth]{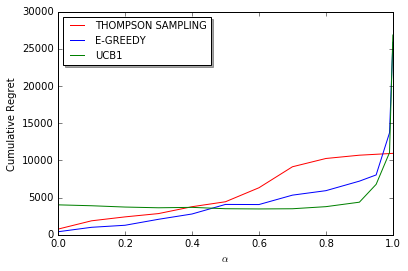}    
    \caption{Change of Cumulative Regret with respect to $\alpha$ where $S=10$, $N=5$ and $\mu=\mu_2$}
\end{figure}

\begin{figure}[H]
  \centering
    \includegraphics[width=0.45\textwidth]{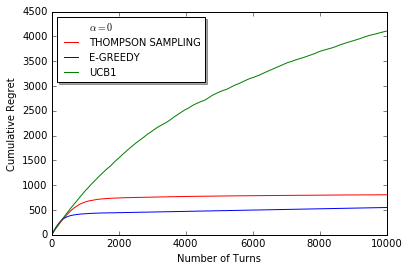}    
    \caption{Change of Cumulative Regret with respect to Number of Turns where $S=10$, $N=5$, $\alpha=0$ and $\mu=\mu_2$}
\end{figure}

\begin{figure}[H]
  \centering
    \includegraphics[width=0.45\textwidth]{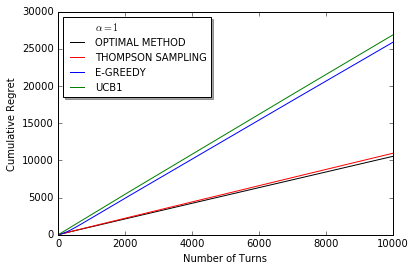}    
    \caption{Change of Cumulative Regret with respect to Number of Turns where $S=10$, $N=5$, $\alpha=1$ and $\mu=\mu_2$}
\end{figure}

\newpage

\section{Conclusion and Future Work}

In this paper, we evaluate a decentralized MAB problem with multiple players in cases of different communication densities between players and using penalty for collisions. Limiting factor in the performance is the collision model. Without collision penalty, the problem can be seen as a single player MAB problem in which pulling multiple arms at the same time is allowed and the only difference than the classic problem is faster convergence to the best arm. We observe that Thompson Sampling usually has the highest performance in terms of minimizing regret among three algorithms where an optimally tuned $\epsilon$-Greedy algorithm can perform best depending on the mean vector $\mu$ of the slots. Also, we conclude that sublinear regret is easily achievable without communication between players, whereas we get linear regret in case of full communication.

Nature of the MAB problem has applications in economics, network communications, bandwidth sharing and game theory where individuals try to maximize their personal utility with limited resources. We perceive this work as a bridge between a classical reinforcement learning problem and game theory in which we analyze different algorithms and test their robustness to communication. We also provide asymptotically optimal strategies for the extreme cases of no communication and full communication.

For future work, optimal strategies for the case $0<\alpha<1$ will be analyzed as it is still unclear how to propose an optimal strategy for any $\alpha$ value. Apart from this, we will evaluate the effect of communication in adversarial and markovian bandits.

\ifCLASSOPTIONcaptionsoff
  \newpage
\fi

\balance

\bibliographystyle{IEEEtran}  
\bibliography{references}

\end{document}